\documentclass[journal]{IEEEtran}
\usepackage{cite}
\usepackage{amsmath,amssymb,amsfonts}
\usepackage{algorithmic}
\usepackage{graphicx}
\usepackage{textcomp}
\usepackage{booktabs}
\usepackage{bm}
\usepackage{array}
\usepackage{hyperref}
\hypersetup{
    colorlinks=true,      
    linkcolor=blue,       
    citecolor=blue,       
    urlcolor=black,       
}

\def\BibTeX{{\rm B\kern-.05em{\sc i\kern-.025em b}\kern-.08em
    T\kern-.1667em\lower.7ex\hbox{E}\kern-.125emX}}

\begin{document}

\title{Hardware-Oriented Inference Complexity of Kolmogorov–Arnold Networks}

\author{Bilal~Khalid,
        Pedro~Freire, 
        Sergei~K.~Turitsyn, 
        and~Jaroslaw~E.~Prilepsky
\thanks{This work has been accepted for publication in IEEE Access. The final published version will be available via IEEE Xplore. This research was supported by the European Union's Horizon Europe research and innovation programme MSCA-DN NESTOR (G.A. 101119983), and by UK Research and Innovation (UKRI) under the Horizon Europe Guarantee scheme (Grant Ref: EP/Y031024/1). Sergei K. Turitsyn also acknowledges EPSRC project TRANSNET (EP/R035342/1). B. Khalid, P. Freire, S. K. Turitsyn, and J. E. Prilepsky are with the Aston Institute of Photonic Technologies, Aston University, Birmingham, B4 7ET, UK. Corresponding author: Pedro Freire (e-mail: p.freiredecarvalhosourza@aston.ac.uk).}}

\markboth{Bilal Khalid \MakeLowercase{\textit{et al.}}: Hardware-Oriented Inference Complexity of Kolmogorov–Arnold Networks}%
{Bilal Khalid \MakeLowercase{\textit{et al.}}: Hardware-Oriented Inference Complexity of Kolmogorov–Arnold Networks}

\maketitle

\begin{abstract}
Kolmogorov-Arnold Networks (KANs) have recently emerged as a powerful architecture for various machine learning applications. However, their unique structure raises significant concerns regarding their computational overhead. Existing studies primarily evaluate KAN complexity in terms of Floating-Point Operations (FLOPs) required for GPU-based training and inference. However, in many latency-sensitive and power-constrained deployment scenarios, such as neural network-driven non-linearity mitigation in optical communications or channel state estimation in wireless communications, training is performed offline and dedicated hardware accelerators are preferred over GPUs for inference. Recent hardware implementation studies report KAN complexity using platform-specific resource consumption metrics, such as Look-Up Tables, Flip-Flops, and Block RAMs. However, these metrics require a full hardware design and synthesis stage that limits their utility for early-stage architectural decisions and cross-platform comparisons. To address this, we derive generalized, platform-independent formulae for evaluating the hardware inference complexity of KANs in terms of Real Multiplications (RM), Bit Operations (BOP), and Number of Additions and Bit-Shifts (NABS). We extend our analysis across multiple KAN variants, including B-spline, Gaussian Radial Basis Function (GRBF), Chebyshev, and Fourier KANs. The proposed metrics can be computed directly from the network structure and enable a fair and straightforward inference complexity comparison between KAN and other neural network architectures. 
\end{abstract}


\section{Introduction}
\label{sec:introduction}

Multi-Layer Perceptrons (MLPs) are the foundational blocks of modern deep learning models and have long served as the universal approximators for non-linear functions~\cite{hornik1989multilayer}. Recently, Kolmogorov-Arnold Networks (KANs)~\cite{liu2025kan} have been proposed as a viable alternative to MLPs. Instead of using fixed activation functions on the network nodes, they incorporate learnable activation functions on the edges. In the classic KAN architecture, these learnable functions are parameterized as B-splines. This deviation from the conventional neural network (NN) architecture allows for more flexible and precise function approximation and offers a higher degree of interpretability~\cite{somvanshi2025survey}. KANs have rapidly gained traction across various domains, including computer vision~\cite{li2025u, abd2024ckan, cheon2024demonstrating}, time-series forecasting~\cite{vaca2024kolmogorov, inzirillo2024sigkan}, graph learning~\cite{zhang2024graphkan, kiamari2024gkan}, physics-informed modeling~\cite{wang2025kolmogorov, rigas2024pikan}, and biomedicine~\cite{knottenbelt2025coxkan}. Moreover, their unique architecture naturally aligns with optical computing principles that also makes them a promising candidate for photonic neuromorphic computing~\cite{peng2024photonic, sozos2026photonic}.

Despite these advantages, significant concerns exist regarding the high computational complexity of KANs, which can pose real-world implementation challenges as compared to conventional architectures such as MLPs. Studies have shown that while KANs can achieve superior performance than MLPs on certain tasks, they require more parameters to do so~\cite{ta2024bsrbf, tafckan, zeng2025kan}. Furthermore, some studies highlight that KANs do not consistently outperform MLPs when the total Floating Point Operations (FLOPs) are controlled~\cite{yu2024kan, hou2024kolmogorov}. In addition, several other works also perform FLOPS-based complexity evaluations of their KAN implementations~\cite{baravsin2025exploring, qiu2025powermlp, ta2025prkan}. However, these studies adopt a GPU-centric perspective optimized for maximum Single-Instruction-Multiple-Data (SIMD) throughput and dense tensor multiplications. While appropriate for characterizing GPU-based training and inference, this approach fails to accurately represent complexity for specialized hardware deployments. For many latency-sensitive and power-constrained applications, such as neural-network-driven non-linearity mitigation in optical communication systems or channel state estimation in wireless communication, Field-Programmable Gate Arrays (FPGAs) or Application-Specific Integrated Circuits (ASICs) are used for inference rather than GPUs. The efficiency is directly linked to the actual number of multipliers and adders instantiated in hardware. In such scenarios, training complexity is largely irrelevant since models are trained once offline on powerful servers, and inference complexity is the critical bottleneck that must be carefully considered.

Furthermore, the hardware complexity of KANs can be fundamentally different from the GPU-oriented FLOPs calculations reported in literature~\cite{yu2024kan, hou2024kolmogorov}. B-splines possess the local support property, i.e., for a spline order $k$ and grid size $G$, only $(k+1)$ basis functions are non-zero for any given input~\cite{liu2025kan}. Properly designed hardware implementations can exploit this sparsity by identifying the active interval and computing only the necessary basis functions. This eliminates the dependence on $G$ that dominates reported FLOPs calculations. Early hardware complexity evaluation studies~\cite{le2024exploring} implemented symbolic approximations of trained KAN models in hardware rather than exploiting local support. Recent optimized hardware implementations of KANs~\cite{huang2025hardware1, errabii2026kan, huang2025hardware2, mammadzada2025design} achieve a substantial reduction in resource consumption as compared to traditional deep neural network (DNN) hardware through careful sparsity-aware design choices. However, these studies report complexity using hardware-specific resource consumption metrics such as Block Random Access Memory (BRAM), Flip-Flops (FF), and Look-Up Tables (LUT). To evaluate these metrics, a complete hardware design and synthesis cycle is required, which makes them unsuitable for early-stage research where many architectural variations have to be tested. Furthermore, these metrics do not reflect the underlying algorithmic costs and depend on the target hardware platform and physical implementation strategy, which prevents meaningful cross-platform comparisons.

This paper addresses these challenges by establishing a comprehensive platform-independent hardware inference complexity evaluation framework for KANs. We derive closed-form expressions for three analytical hardware complexity metrics, i.e., Real Multiplications (RM), Bit Operations (BOP), and Number of Additions and Bit Shifts (NABS). In previous work~\cite{freire2024computational}, some of the present authors evaluated these metrics for other DNN architectures. The calculations are now extended to encompass four major KAN variants, namely B-spline, Gaussian Radial Basis Function (GRBF), Chebyshev polynomial, and Fourier basis KANs. Unlike platform-specific resource metrics, these analytical metrics can be computed directly from the network architecture. This enables rapid design-space exploration and fair comparison between KANs and other DNN architectures. The main contributions of this paper can be summarized as follows:

\begin{itemize}
    \item Recent hardware-oriented KAN studies~\cite{le2024exploring, huang2025hardware1, errabii2026kan, huang2025hardware2, mammadzada2025design} implement specific designs and report their cost through platform-specific resource-utilization metrics. In contrast, to the best of our knowledge, this is the first study to characterize the inference complexity of four different KAN variants through platform-independent analytical metrics, namely RM, BOP, and NABS. We derive generalized formulae for evaluating the per-edge and per-layer complexity in terms of these metrics.
    
    \item We highlight that existing FLOPs calculations for KANs are appropriate for GPU-based training and inference but overestimate complexity for sparsity-aware optimized hardware implementations.

    \item We perform a thorough iso-complexity analysis to demonstrate the specific network size reductions required for KAN variants to offset their high per-edge hardware costs when compared to standard MLPs.
        
    \item We provide platform-independent analytical complexity metrics that facilitate the research community to make robust and fair hardware inference complexity comparison of different KAN variants with other DNNs.
\end{itemize}

The remainder of this paper is organized as follows. Section~\ref{sec:background} provides background on KAN architecture and basis function variants. An overview of commonly used metrics for complexity evaluation of KANs is presented in Section~\ref{sec:existing}. A description of the proposed analytical complexity metrics is presented in Section~\ref{sec:proposed} followed by their comprehensive mathematical derivation in Section~\ref{sec:derivation}. A demonstrative comparative analysis of KANs with MLPs in terms of these metrics is performed in Section~\ref{sec:comparison}, followed by conclusion in Section~\ref{sec:conclusion}.

\section{Background}
\label{sec:background}

\subsection{Kolmogorov-Arnold Network (KAN) Architecture}

\begin{figure}[t]
    \centering
    \includegraphics[width=\linewidth]{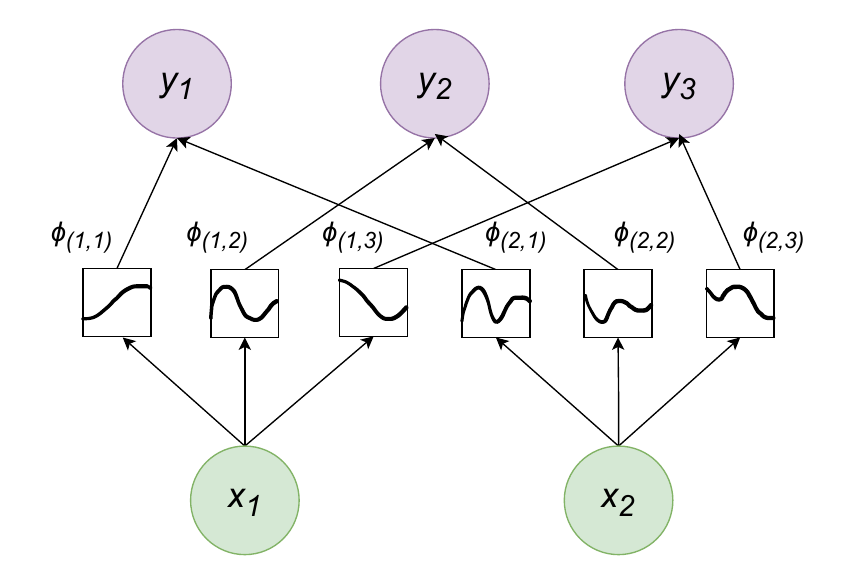}
    \caption{A KAN layer with two input and three output nodes. Unlike MLPs where activations reside at nodes, each KAN edge contains a learnable activation function.}
    
    \label{fig:kan_architecture}
\end{figure}

\begin{figure*}[t]
    \centering    \includegraphics[width=0.85\textwidth, height=0.6\textheight, keepaspectratio]{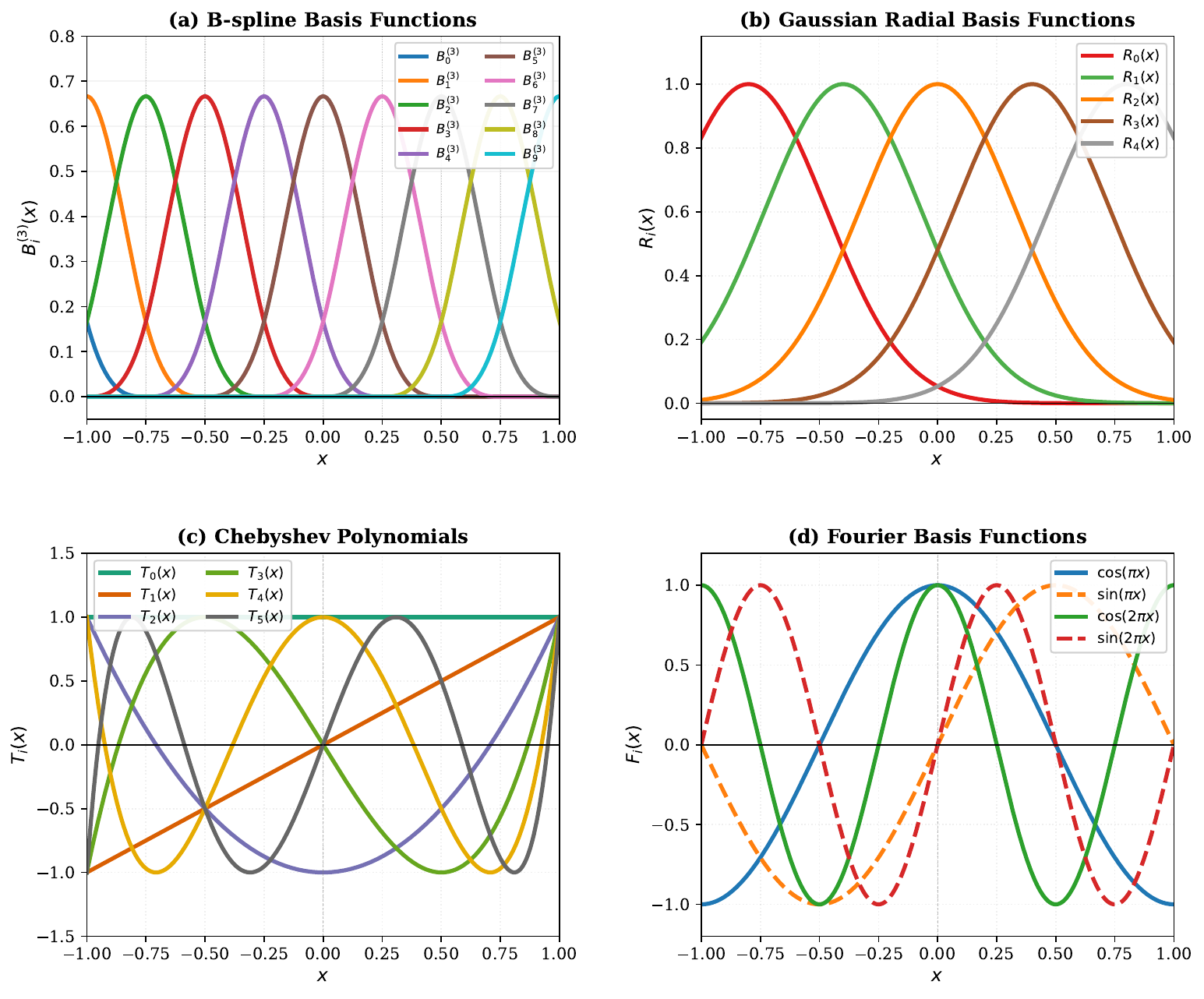}
    \caption{Basis functions commonly used in KANs: (a) B-splines; (b) Gaussian radial basis functions (GRBF); (c) Chebyshev polynomials; (d) Fourier basis.}
    \label{fig:basis_comparison}
\end{figure*}

KANs represent a fundamental departure from the conventional MLP architecture. This distinction is rooted in the Kolmogorov-Arnold representation theorem~\cite{kolmogorov1957representations}, which establishes that any multivariate continuous function can be represented as a finite composition of univariate continuous functions and additions. Figure~\ref{fig:kan_architecture} illustrates the fundamental architecture of a KAN layer, where learnable activation functions are placed on the edges. Extending this structure to a generalized layer with $n_i$ input and $n_n$ output dimensions, the value at each output neuron $y_j$ is computed as
\begin{equation}
    y_j = \sum_{i=1}^{n_i} \phi_{i,j}(x_i), \quad j = 1, \, 2, \, \ldots, \, n_n \, ,
\end{equation}
where $\phi_{i,j}(\cdot)$ represents the learnable activation function connecting input $x_i$ to output $y_j$. Unlike a standard MLP neuron that performs simple linear weighting, a KAN edge learns $\phi(x)$ that is composed of a residual base path and a parameterized basis function representation. This can be expressed as
\begin{equation}
\label{eq:kan_edge_generic}
    \phi(x) = w_b \, \sigma(x) + \sum_{i=1}^{N} w_i \, f_i(x),
\end{equation}
where $\sigma(x)$ is typically the base activation function\footnote{SiLU is commonly used, but theoretically, any standard activation function can be used. The base activation path $w_b \sigma(x)$ facilitates training by providing a globally defined component}, $w_b$ is the learnable base weight, $w_i$ denote the basis control coefficients, $f_i(x)$ represent the basis functions, and $N$ denotes the number of basis functions. Recently, numerous architectural variations of KANs have emerged that incorporate different basis functions~\cite{noorizadegan2025practitioner}. These include the originally proposed B-splines~\cite{liu2025kan, liu2024kan2}, Gaussian radial basis functions (GRBF)~\cite{ta2024bsrbf,abueidda2025deepokan,li2024kolmogorov}, Chebyshev polynomials~\cite{ss2024chebyshev, torchkan}, Fourier series~\cite{xu2024fourierkan} and several others~\cite{aghaei2025fkan, bozorgasl2024wavkanwaveletkolmogorovarnoldnetworks, seydi2024unveiling}. The choice of basis function fundamentally determines the expressiveness, approximation properties, and computational characteristics of the KAN architecture. In this paper, we restrict our complexity analysis to four representative types, namely B-splines, GRBFs, Chebyshev polynomials, and Fourier series. Figure~\ref{fig:basis_comparison} illustrates the characteristic shapes of these basis functions. The mathematical formulation of each type and its implications for hardware complexity are discussed next.

\subsection{B-spline Basis Functions}

B-splines are piecewise polynomial functions widely used in numerical analysis and approximation theory~\cite{de1978practical}. In the context of KANs, B-splines serve as the parametric representation for learnable activation functions. A B-spline of order $k$ is defined over a knot vector, which partitions the input domain into intervals. Let $\mathcal{T} = \{t_0, t_1, \ldots, t_{G}\}$ denote a non-decreasing sequence of $(G+1)$ knots that divides the input range $[a, b]$ into $G$ intervals. To maintain continuity at the boundaries, $\mathcal{T}$ is extended by adding $k$ additional knots at each end such that  $\mathcal{T} = \{t_{-k}, \dots, t_{-1}, t_0, t_1, \dots, t_G, t_{G+1}, \dots, t_{G+k}\}$ resulting in a total of $(G+2k+1)$ knots. A B-spline basis function $B_{i,k}(x)$ is defined recursively using the Cox-de Boor formula~\cite{de1978practical}:
\begin{equation}
\label{eq:cox_deboor_recursive}
    B_{i,k}(x) = \frac{x - t_i}{t_{i+k} - t_i} B_{i, k-1}(x) + \frac{t_{i+k+1} - x}{t_{i+k+1} - t_{i+1}} B_{i+1, k-1}(x),
\end{equation}
\begin{equation}
\label{eq:cox_deboor_base}
    B_{i,0}(x) = 
    \begin{cases} 
        1 & \text{if } t_i \le x < t_{i+1}, \\
        0 & \text{otherwise},
    \end{cases}
\end{equation}
where $B_{i,0}(x)$ is the zero-order basis function used to recursively build higher-order basis functions. For B-spline KANs, the edge function in Eq.~\eqref{eq:kan_edge_generic} becomes:
\begin{equation}
\label{eq:kan_edge_bspline}
    \phi(x) = w_b \, \sigma(x) + \sum_{i=0}^{G+k-1} w_i \, B_{i,k}(x).
\end{equation}

The most critical property of B-splines for hardware implementation is their local support. A $k^{th}$-order B-spline basis function $B_{i,k}(x)$ is non-zero only in the interval $[t_{-k+i}, t_{i+1}]$. Consequently, for any input value $x$, only $(k+1)$ basis functions are non-zero. To illustrate, consider a cubic B-spline ($k=3$) defined over a grid with $G=50$ intervals. Although the total number of basis functions is $(G+k) = 53$, for any specific input $x$, only $4$ of these basis functions are non-zero. This implies that the summation in Eq.~\eqref{eq:kan_edge_bspline} reduces from a dense operation over all $53$ terms to a sparse operation over just $4$ terms. This significantly reduces the computational cost as hardware resources scale with $k$, typically a small constant, rather than with $G$, which may be large to achieve high approximation fidelity.

\subsection{Gaussian Radial Basis Functions (GRBF)}

Gaussian Radial Basis Functions (GRBFs) provide an alternative option to B-splines. GRBF KANs were introduced to mitigate the complex grid management and computational overhead associated with B-spline KANs~\cite{ta2024bsrbf,abueidda2025deepokan}. They simplify the underlying architecture and accelerate the processing by replacing the B-spline expansion with a sum of Gaussian functions. Each GRBF is defined as
\begin{equation}
\label{eq:grbf}
    R_i(x) = \exp\left(-\frac{(x - c_i)^2}{2\sigma_i^2}\right),
\end{equation}
where $c_i$ is the center of the $i$-th Gaussian and $\sigma_i$ is the width parameter. The edge function for GRBF-KAN can therefore be written as
\begin{equation}
\label{eq:kan_edge_grbf}
    \phi(x) = w_b \, \sigma(x) + \sum_{i=0}^{N_c-1} w_i \, R_i(x),
\end{equation}
where $N_c$ denotes the number of Gaussian centers. Each GBRF is non-zero over the entire input domain. In principle, all $N_c$ basis functions need to be evaluated for every input. However, by adopting a constant width parameter $\sigma$ and fixing the grid size, all GRBFs become shifted versions of each other. As a result, only the weights remain edge-specific, which enables the reuse of the basis functions.

\subsection{Chebyshev Polynomial Basis}

Chebyshev polynomials provide a classical orthogonal polynomial basis for KANs. By using an orthogonal basis, Chebyshev KANs ensure smooth and stable function approximation across the entire input domain~\cite{ss2024chebyshev}. Chebyshev polynomials of the first kind are defined by the three-term recurrence relation given as
\begin{equation}
\label{eq:chebyshev_recurrence}
    T_i(u) = 2u \, T_{i-1}(u) - T_{i-2}(u),
\end{equation}
with base terms $T_0(u) = 1$ and $T_1(u) = u$, where $u \in [-1, 1]$. Since Chebyshev polynomials are properly defined on the interval $[-1, 1]$, the input is typically normalized using $u = \tanh(x)$ to map the unbounded input domain to the valid range. The edge function for Chebyshev KAN can therefore be written as
\begin{equation}
\label{eq:kan_edge_chebyshev}
    \phi(x) = w_b \, \sigma(x) + \sum_{i=0}^{n} w_i \, T_i(\tanh(x)),
\end{equation}
where $n$ is the maximum polynomial degree. Chebyshev polynomials are globally defined over $[-1, 1]$ with no local support property. All $(n+1)$ polynomial terms must be evaluated for every input. This causes the computational cost to scale linearly with the polynomial degree $n$. However, unlike B-splines where the total number of basis functions $(G+k)$ can be large, Chebyshev KANs typically use relatively low polynomial degrees due to the global nature of orthogonal polynomials.

\subsection{Fourier Series Basis}

Fourier-based KANs~\cite{xu2024fourierkan} decompose activation functions into frequency components using trigonometric basis functions. They are particularly well-suited for learning periodic or oscillatory functions, which can make them an attractive choice for signal processing applications. The KAN edge function can be expressed in terms of a Fourier series as
\begin{equation}
\label{eq:kan_edge_fourier}
    \phi(x) = w_b \, \sigma(x) + \sum_{i=1}^{G} \left[a_i \cos(i\omega x) + b_i \sin(i\omega x)\right],
\end{equation}
where $G$ is the grid parameter that controls the number of frequency components, $\omega$ is the angular frequency, and $a_k, b_k$ are the learnable Fourier coefficients. The Fourier basis consists of $2G$ functions, i.e., $G$ cosine terms and $G$ sine terms. Like Chebyshev polynomials, Fourier basis functions are globally defined. Each cosine and sine component spans the entire input domain, which requires all $2G$ basis functions to be evaluated for every input. 

\section{Metrics for Complexity Evaluation of KANs}
\label{sec:existing}
This section briefly discusses the main metrics that have been used in the literature to evaluate the computational complexity of KANs.

\subsection{Number of Parameters (\texorpdfstring{$N_{par}$}{Npar})}

$N_{par}$ quantifies the total number of learnable weights in a model. It is a measure of the memory required to store the trained network and serves as an indicator of model capacity. The learnable parameters in B-spline KANs include the spline control points, the shortcut connection weights, the spline weights, and the bias. The total learnable parameters per layer are given as~\cite{yu2024kan}:
\begin{equation}
    N_{par} = (n_i \times n_n) \times (G + k + 3) + n_n. 
\end{equation}
Thus, if $N$ denotes the width of the layers and $L$ denotes the depth of the network, the memory complexity in terms of parameters is given by $O(N^2 L (G+k))$~\cite{liu2025kan}. The $N^2$ term reflects the fully-connected nature of the layers, while the $(G+k)$ term represents the additional parameters required for every edge. In contrast, an MLP of the same width and depth only needs $O(N^2L)$ parameters.

\subsection{Floating Point Operations (FLOPs)}

FLOPs represent the total number of mathematical operations required to execute a single forward pass. A multiply-and-accumulate (MAC) operation is counted as two FLOPs i.e. one multiplication and one addition. The FLOPs required to evaluate the B-splines in a single KAN layer are expressed as~\cite{yu2024kan, hou2024kolmogorov}:
\begin{equation}
\label{eq:flops_layer}
\begin{split}
    &FLOPS_{Layer} = (n_i \times n_n) \\ &\times \bigl[ 9\times k \times \,(G + 1.5 \times k) 
    + 2 \times G - 2.5 \times k - 1 \bigr].
\end{split}
\end{equation}
The above calculation considers dense tensor operations where all $(G+k)$ B-spline basis functions need to be computed. This approach is suitable for evaluating training complexity because gradients have to be updated across all of the control points. It is also appropriate for GPU-based inference, as GPUs are optimized for dense matrix multiplications. In such cases, maintaining parallelization is often more efficient than using sparse indexing to identify active basis functions. However, as already discussed, GPUs are often unsuitable for inference tasks in power-constrained and latency-sensitive environments due to their high cost and energy consumption. Efficient hardware implementations can take advantage of the B-spline sparsity and compute only the active $(k+1)$ basis functions. Consider a single cubic B-spline edge defined over a grid of $G=50$ intervals. Evaluating the FLOPs of a single edge using Eq.~\eqref{eq:flops_layer} gives $1563$ FLOPs, dominated by the $9kG$ term that scales linearly with the grid size. A sparsity-aware hardware implementation instead evaluates only the $k+1=4$ active basis terms independent of $G$. Hence, a FLOPs calculation that scales with $G$ significantly overestimates the inference complexity of a B-spline KAN for an efficient hardware implementation.

\subsection{Hardware Resource Consumption Metrics}

In hardware implementations of KAN~\cite{le2024exploring, errabii2026kan, huang2025hardware1, huang2025hardware2, mammadzada2025design}, the complexity is evaluated using platform-specific resource utilization metrics. Unlike analytical metrics that can be computed directly from the network architecture, these metrics require a hardware design and synthesis stage. This limits their utility for early-stage architectural exploration or cross-platform comparisons. The four primary resource metrics reported by FPGA design tools are Block RAMs (BRAMs), Digital Signal Processing blocks (DSPs), Look-Up Tables (LUTs) and Flip-Flops (FFs). In addition, on-chip memory footprint and occupied silicon area can also be considered as indicators of complexity. 

\section{Analytical Hardware Inference Complexity Metrics for KANs}
\label{sec:proposed}

In Ref.~\cite{freire2024computational}, some of the present authors analyzed the complexity of various NN architectures in terms of RM, BOP, and NABS. Although these metrics are commonly used for assessing the computational complexity of other NNs, so far, a comprehensive evaluation of KANs in terms of these metrics has not been presented. We briefly detail what these metrics represent below.

\subsection{Real Multiplications (RM)}

RM counts only the number of multiplications required by an algorithm. It ignores additions because multiplication operations are significantly more expensive in hardware than additions. They are also slower and consume more chip area~\cite{mirzaei2006fpga}. For floating-point arithmetic with fixed precision, multiplication dominates the computational cost. Therefore, RM provides a useful first-order comparison between different architectures when implemented with the same numerical precision.

\subsection{Bit Operations (BOP)}

The hardware complexity of quantized NNs can be evaluated using the BOP metric. Unlike RM, BOP accounts for the bitwidth precision of operands and distinguishes between the costs of multiplication and accumulation. It is suitable for fixed-point and mixed-precision implementations where different operations may use different bitwidths. The cost of multiplying two operands scales with the product of their bitwidths, whereas the accumulation cost scales with the accumulator bitwidth. 

\subsection{Number of Additions and Bit Shifts (NABS)}

NABS accounts for the fact that fixed-point multiplications can also be implemented using bit-shifters and adders. Since bit shifts incur negligible hardware cost as compared to additions, the NABS metric represents the complexity solely in terms of the equivalent number of adders required. To calculate NABS for KAN, we replace each multiplication in the BOP formulation with $X$ number of adders. The value of $X$ depends on the quantization scheme. For uniform quantization, $X = b - 1$ where $b$ denotes the bitwidth. Typically, $X = 0$ for Power-of-Two (PoT) quantization~\cite{przewlocka2022power} and $X = n$ for Additive Powers-of-Two (APoT) quantization~\cite{Li2020Additive} with $n$ additive terms.

\section{Evaluation of Metrics for different KANs}
\label{sec:derivation}

In this section, we derive the analytical expressions for inference complexity metrics for B-spline, GRBF, Chebyshev, and Fourier KANs. We analyze the inference complexity of each variant in terms of RM, BOP, and NABS. We evaluate the complexity per edge and then generalize the formulae for dense KAN layers. For these calculations, we assume that LUTs are used for the implementation of fixed activation functions, such as those applied to nodes in MLPs or to the residual base path in KANs, as is typically the case~\cite{freire2024computational}.

\subsection{B-Spline KANs}
We first derive the generalized formula for Real Multiplications per edge $(RM_{edge})$ for B-spline KANs. As shown in Eq.~\eqref{eq:kan_edge_bspline}, a B-spline KAN edge learns a non-linear activation function $\phi(x)$ composed of a residual base path and a parameterized B-spline path.  We define $RM_{edge}$ as the sum of three structural components i.e., fixed overhead $(C_{fixed}$), linear combination cost $(C_{lin})$, and basis evaluation cost $(C_{basis})$, such that
\begin{equation}
    RM_{edge} = C_{fixed} + C_{lin} + C_{basis}.
\end{equation}
The first term, $C_{fixed}$, represents the cost of required static operations. For B-spline KANs, the input $x$ is normalized to the grid index that requires one multiplication. Additionally, the residual base path term $w_b \sigma(x)$ also requires one multiplication to scale the non-linear activation by the learnable weight. Therefore, the fixed overhead in terms of multiplications can be represented as $C_{fixed_{(bspline)}}=2$. The second component, $C_{lin}$, arises from the weighted summation term in Eq.~\eqref{eq:kan_edge_bspline}. Due to the local support property of B-splines, only $(k+1)$ basis functions are non-zero within any specific grid interval. This reduces the global summation to a localized dot product between these $(k+1)$ active basis values and their corresponding control coefficients. Hence, the linear combination cost in terms of RM is given as $C_{lin_{(bspline)}} = k + 1$.

The third component, $C_{basis}$, accounts for the computational cost of generating the basis function values $B_{i,k}(x)$ themselves. These can be computed via the Cox-de Boor recursion, Eq. ~\eqref{eq:cox_deboor_recursive} and Eq.~\eqref{eq:cox_deboor_base}. The recursion constructs higher-order basis functions as a weighted linear combination of two lower-order basis functions. The base case $B_{i,0}(x)$ returns $1$ if the input $x$ falls within the specific grid interval $[t_i, t_{i+1}]$, and $0$ otherwise. We consider a hardware-optimized implementation that recognizes boundary conditions in the recursive triangle~\cite{Piegl1997}. At any recursive depth $p$, where $2 \le p \le k$, there are $(p+1)$ active nodes. The first and last nodes ($B_{0,p}, B_{p,p}$) have only one non-zero parent terms since they are at the edges of the recursive triangle. Thus, they require only $1$ multiplication each. The remaining $(p-1)$ internal nodes mix two non-zero parent terms, requiring $2$ multiplications each. Thus, at depth $p$, the multiplication cost $(C_p)$ is given by
\begin{equation}
    \text{C}_p = (2 \times 1) + (p-1) \times 2 = 2p.
\end{equation}
By adding this cost from depth $p=2$ to $k$, we obtain the total RM cost of evaluating the basis functions per edge as
\begin{align}
    C_{basis_{(bspline)}} &= \sum_{p=2}^{k} C_p =\sum_{p=2}^{k} (2p)\notag \\ 
    &= 2 \left[\sum_{p=1}^{k} (p) - 1 \right] =2 \left[ \frac{k(k+1)}{2} - 1 \right]\notag \\
    &= k^2 + k - 2.
\end{align}
Thus, $RM_{edge_{(bspline)}}$ can be obtained by the summation of the three component terms i.e.
\begin{align}
    RM_{edge_{(bspline)}} &=  C_{fixed_{(bspline)}} + C_{lin_{(bspline)}} + C_{basis_{(bspline)}} \notag \\
    &= 2 + (k + 1) + (k^2 + k - 2) \notag \\
    &= (k+1)^2.
\end{align} 
However, this RM cost can be further reduced through highly optimized hardware implementations as demonstrated in~\cite{errabii2026kan}. The recursive calculation can be replaced with a tabulation strategy that relies on the inherent translation and scaling invariance of B-splines. It essentially reduces the evaluation of any basis function $B_{i,k}(x)$ to a simple lookup from Read-Only Memory (ROM). This transition from mathematical recursion to a memory-based lookup effectively eliminates the high-cost of recursive multiplications associated with the Cox-de Boor formula. Consequently, the basis evaluation cost ($C_{basis_{(bspline)}}$) in terms of RM is effectively reduced to zero. Thus, for optimized hardware implementations, $(RM_{edge_{(bspline)}})$ can be recalculated as
\begin{align}
    RM_{edge_{(bspline)}} &=  C_{fixed_{(bspline)}} + C_{lin_{(bspline)}} \notag \\
    &= k + 3.
\end{align}
For a dense layer with $n_i$ input nodes and $n_n$ output nodes, the total RM per layer $RM_{layer_{(bspline)}}$ can be written as
\begin{equation}
    RM_{Layer_{(bspline)}} = n_n n_i (k + 3).
\end{equation}

\begin{table*}[t]
\centering
\small
\setlength{\tabcolsep}{5pt}
\renewcommand{\arraystretch}{1.6}
\caption{All formulae represent complexity per-layer with $n_i$ inputs and $n_n$ outputs. $M$ denotes the number of active basis terms per edge. The per-edge terms, $\mathrm{BOP}_{edge~}$ and $\mathrm{NABS}_{edge}$, are given in full by the corresponding equations in Section~\ref{sec:derivation}; $b$ denotes the basis bitwidth and $*$ represents the relevant basis function (B-splines, GRBF, Chebyshev or Fourier). This table was condensed during revision to improve readability.}
\label{tab:complexity_all_networks}
\begin{tabular}{|l|c|c|c|c|}
\hline
\textbf{Network} & \textbf{M} & \textbf{RM / layer} & \textbf{BOP / layer} & \textbf{NABS / layer} 
\\
\hline\hline
MLP & --- & $n_n n_i$ & $n_n n_i\,[\,b_w b_i + \text{Acc}(n_i,b_w,b_i)\,]$ & $n_n n_i (X_w+1)\,\text{Acc}(n_i,b_w,b_i)$ \\
\hline
B-spline & $k{+}1$ & $n_n n_i (k+3)$ & $\mathcal{B}_{bspline}$ & $\mathcal{N}_{bspline}$ \\
\hline
GRBF & $N_c$ & $n_n n_i (N_c+1)$ & $\mathcal{B}_{rbf}$ & $\mathcal{N}_{rbf}$ \\
\hline
Chebyshev & $n{+}1$ & $n_n n_i (n+2)$ & $\mathcal{B}_{cheby}$ & $\mathcal{N}_{cheby}$ \\
\hline
Fourier & $2G$ & $n_n n_i (2G+1)$ & $\mathcal{B}_{fourier}$ & $\mathcal{N}_{fourier}$ \\
\hline
\multicolumn{5}{|l|}{The per-layer BOP and NABS formulae for each KAN variant follow a similar structure:} \\
\multicolumn{5}{|l|}{$\mathcal{B}_*=n_n n_i\,\mathrm{BOP}_{edge*} + n_n (n_i-1)\big[\text{Acc}(M_*,b_w,b_{\star})+\lceil\log_2 n_i\rceil\big]$} \\
\multicolumn{5}{|l|}{$\mathcal{N}_*=n_n n_i\,\mathrm{NABS}_{edge*} + n_n (n_i-1)\big[\text{Acc}(M_*,b_w,b_{\star})+\lceil\log_2 n_i\rceil\big]$} \\
\hline
\end{tabular}
\end{table*}

We then evaluate the BOP metric following the methodology adopted in~\cite{freire2024computational}. For a standard dense MLP layer, the BOP is composed of the operations required for the vector-matrix multiplication ($BOP_{Mul}$) and the bias addition ($BOP_{Bias}$). For a single MLP neuron with $n_i$ inputs, each multiplication involving a weight of bitwidth $b_w$ and an input of bitwidth $b_i$ requires $(n_i b_w b_i)$ bit operations. Moreover, summing these $n_i$ products requires $(n_i - 1)$ additions. To prevent overflow, the accumulator must be able to accommodate the product size plus the additional growth resulting from the summation. Therefore, $BOP_{Mul}$ and $BOP_{Bias}$ for a dense MLP layer are calculated as
\begin{equation}
\begin{split}
    BOP_{Mul} = n_n \bigl[n_i b_w b_i + (n_i - 1) \\
    \times (b_w + b_i + \lceil \log_2(n_i) \rceil)\bigr],
\end{split}
\end{equation}
\begin{equation}
    BOP_{Bias} = n_n \big[(b_w + b_i + \lceil \log_2(n_i) \rceil)\big].
\end{equation}
To be consistent with~\cite{freire2024computational}, we also adopt the short notation $Acc()$ that represents the accumulator bitwidth needed for the multiply and accumulate (MAC) operation. We define
\begin{equation}
\label{eq:accumulator}
    \text{Acc}(n_i, b_w, b_i) = b_w + b_i + \lceil \log_2(n_i) \rceil.
\end{equation}
Equation~\eqref{eq:accumulator} specifies the \emph{overflow-free} accumulator width. It combines the product width $(b_w + b_i)$ with $\lceil \log_2 n_i \rceil$ guard bits. These guard bits accommodate the worst-case bit growth when summing $n_i$ products such that no saturation is required. However, in resource-limited hardware designs, the accumulator is instead clamped to a fixed maximum width, $b_{acc}$, evaluated as
\begin{equation}
\label{eq:accumulator_sat}
    \text{Acc}_{sat}(n_i, b_w, b_i, b_{acc}) = \min\!\big(b_{acc},\; b_w + b_i + \lceil \log_2(n_i) \rceil \big).
\end{equation}
This clamping operation introduces a per-chain saturation cost, $C_{sat} \approx b_{acc}$, to account for the necessary comparison and bounding logic. Furthermore, rounding the result to a smaller output width introduces a rounding cost, $C_{rnd} \approx b_{acc}$. If the result is simply truncated instead of rounded, $C_{rnd} = 0$. Critically, both $C_{sat}$ and $C_{rnd}$ scale linearly as $O(b_{acc})$ per output node. In contrast, the multiplier array cost scales quadratically as $O(b^2)$. Since these addition-based operations are sub-leading, they are omitted from the primary complexity metrics reported in this study. Nevertheless, these can be reinstated as an additive term $n_n(C_{sat} + C_{rnd})$ whenever exact bit-accurate accounting is required. Thus, the total BOP complexity of a dense MLP layer with $n_n$ nodes can be represented by the follwing expression:
\begin{align}
    BOP_{Layer_{MLP}} &=  BOP_{Mul} + BOP_{Bias} \notag \\
    \begin{split}
        &= n_n \bigl[n_i b_w b_i + (n_i - 1)(b_w + b_i + \lceil \log_2(n_i) \rceil) \\
        &\quad + (b_w + b_i + \lceil \log_2(n_i) \rceil)\bigr] \notag
    \end{split} \\
    &= n_nn_i \left[b_w b_i + (b_w + b_i + \lceil \log_2(n_i) \rceil)\right] \notag \\
    &= n_nn_i \left[b_w b_i + \text{Acc}(n_i, b_w, b_i)\right].
\end{align}
The BOP calculation for a single KAN edge can be performed following a similar approach. As with the RM calculation, the fixed overhead accounts for the initial input normalization and the base activation path. Mapping the input of bitwidth $b_i$ to the grid requires a subtraction $(b_i$~BOPs) and a multiplication by the grid reciprocal. If $b_{knot}$ denotes the bitwidth precision of the grid spacing, this multiplication cost can be calculated as $b_ib_{knot}$ BOPs. Moreover, the fixed cost of the multiplication between base weight and the base activation $(w_b \sigma(x))$ can be represented as $b_ib_w$ BOPs. Therefore, we have
\begin{align}
\label{eq:BOP_fixed_kan}
    BOP_{fixed_{(bspline)}} &= b_i + b_ib_{knot} + b_ib_w \notag\\
    &= b_i (1 + b_{knot} + b_w).
\end{align}
In the optimized hardware implementation, the computational burden of basis evaluation is eliminated. Since the basis function values are obtained via memory fetches rather than arithmetic MAC operations, the BOP contribution can be ignored i.e. $BOP_{basis_{(bspline)}} \approx 0$. Finally, the cost of the weighted summation of the active basis functions $(\sum w_i B_i(x))$ has to be considered. Since only $(k+1)$ basis functions are non-zero, this operation is a dot product of size $(k+1)$, involving weights of bitwidth $b_w$ and basis values of bitwidth $b_{basis}$. Thus, the BOP cost for this linear combination can be calculated as
\begin{align}
\label{eq:BOP_linear_kan}
    BOP_{lin_{(bspline)}} &= (k+1) b_w b_{basis} \notag \\
    &\quad + k(b_w + b_{basis} + \lceil \log_2(k+1) \rceil) \notag \\
    &= (k+1) b_w b_{basis} \notag \\
    &\quad + k \left[\text{Acc}(k+1, b_w, b_{basis})\right].
\end{align}
We can therefore obtain the total bit operations for a hardware-optimized KAN edge by adding the cost of individual components i.e.
\begin{align}
    BOP_{edge_{(bspline)}} &= BOP_{fixed_{(bspline)}} + BOP_{lin_{(bspline)}} \notag \\
    \begin{split}
        &= b_i (1 + b_{knot} + b_w) + (k+1) b_w b_{basis} \\
        &\quad + k\left[\text{Acc}(k+1, b_w, b_{basis})\right].
    \end{split}
\end{align}
To derive the total complexity for a dense KAN layer with $n_i$ input and $n_n$ output neurons, we also have to account for the final accumulation cost at each output node. The bitwidth of the accumulation result must be large enough to prevent overflow. The maximum bitwidth of a B-spline KAN edge is given by $\text{Acc}(k+1, b_w, b_{basis})$. The accumulator at each of the $n_n$ output nodes, where $n_i$ edges are summed, has to handle further additional bit growth. Hence, the final BOP calculation for a dense B-spline KAN layer is expressed as
\begin{multline}
\label{eq:BOP_layer_kan}
    BOP_{Layer_{(bspline)}} = n_n n_i [BOP_{edge_{(bspline)}}] \\
    + n_n \left[ (n_i - 1)(\text{Acc}(k+1, b_w, b_{basis}) + \lceil \log_2 n_i \rceil) \right].
\end{multline}

To calculate the NABS, we replace each multiplication in the BOP calculation with $X_w$ adders. For the fixed overhead derived in Eq.~\eqref{eq:BOP_fixed_kan}, the subtraction cost remains unchanged since no multiplication is involved. The grid normalization and base path multiplications are replaced with equivalent additions that gives
\begin{equation}
    NABS_{fixed_{(bspline)}} = b_i + X_{knot}(b_i + b_{knot}) + X_w(b_i + b_w),
\end{equation}
where $X_{knot}$ and $X_w$ represent the maximum number of adders required to perform the multiplication for the grid and weight parameters, respectively. For the linear combination component derived in Eq.~\eqref{eq:BOP_linear_kan}, the $(k+1)$ multiplications are substituted with equivalent additions whereas the $k$ accumulations remain unchanged. Thus, the NABS for the linear combination are given by
\begin{multline}
    NABS_{lin_{(bspline)}} = (k+1) X_w \left[ \text{Acc}(k+1, b_w, b_{basis})\right] \\
    + k \left[ \text{Acc}(k+1, b_w, b_{basis})\right] \\
    =\left[(k+1)X_w + k\right] \left[ \text{Acc}(k+1, b_w, b_{basis})\right].
\end{multline}
The total NABS per KAN edge can therefore be calculated as
\begin{multline}
    NABS_{edge_{(bspline)}} = NABS_{fixed_{(bspline)}} + NABS_{lin_{(bspline)}} \\
    = b_i + X_{knot}(b_i + b_{knot}) + X_w(b_i + b_w) \\
    + \left[(k+1)X_w + k\right] \left[\text{Acc}(k+1, b_w, b_{basis})\right].
\end{multline}
Finally, the NABS for a dense B-spline KAN layer sums the edge costs across all $n_i \times n_n$ connections and includes the final output accumulation. Since the output accumulation consists entirely of additions, it carries over directly from the BOP formulation shown in Eq.~\eqref{eq:BOP_layer_kan}:
\begin{multline}
    NABS_{Layer_{(bspline)}} = n_n n_i \left[NABS_{edge_{(bspline)}}\right] \\
    + n_n \left[(n_i - 1)\left(\text{Acc}(k+1, b_w, b_{basis}) + \lceil \log_2 n_i \rceil\right)\right].
\end{multline}

\subsection{Gaussian Radial Basis Function (GRBF) KANs}

We calculate the complexity metrics for GRBF KANs following the same methodology that was adopted for B-Spline KANs. For GRBF KANs, two cases can be considered. In the first case, the hardware calculates the GRBFs for every forward pass. This requires instantiating multipliers and adders to handle the distance calculation, squaring, scaling, and the final weighted sum. In the second, GRBFs are precomputed and stored in memory. If $N_c$ is fixed, only $N_c$ LUTs are needed for the entire network. Moreover, if the shape of each GRBF is also fixed, i.e. fixed $\sigma$, only $1$ LUT has to be saved since all GRBFs essentially become shifted versions of each other. This is similar to the case for optimized B-spline KAN implementations, where only the cardinal B-spline needs to be saved due to the translation and scaling invariance~\cite{errabii2026kan}.

We follow the decomposition strategy adopted for B-spline KANs for all other variants where the total complexity is considered as the sum of fixed, basis evaluation and linear combination costs. For the RM calculation, the fixed overhead consists of only the base path multiplication $(w_b \sigma(x))$. For GRBF KANs with $N_c$ separate LUTs and fixed centers, LUTs can be indexed by the input directly and no multiplication is necessary for normalization. Thus, we have $C_{fixed_{(GRBF)}} = 1$. Moreover, the linear combination cost considers the weighted sum $\sum_{i=0}^{N_c-1} w_i R_i(x)$ which requires $N_c$ multiplications. Therefore, we define $C_{lin_{(GRBF)}} = N_c$.
Furthermore, for the GRBF evaluation, a multiplication is required for squaring the distance from the center $(x - c_i)^2$ and another for scaling by the constant $-1 / 2\sigma^2$. We ignore the cost of the exponential calculation, since LUTs are commonly used to compute exponentials in hardware. Thus, we consider $2$ multiplications for each center, i.e. $C_{basis_{(GRBF)}} = 2N_c$. Therefore, $RM_{edge_{(GRBF)}}$ can be calculated as
\begin{align}
    RM_{edge_{(GRBF)}} &= C_{fixed_{(GRBF)}} + C_{lin_{(GRBF)}} + C_{basis_{(GRBF)}} \notag \\
    &= 3N_c + 1.
\end{align}
However, with LUT-based optimization, each $R_i(x)$ can be retrieved via memory access, which essentially makes $C_{basis_{(GRBF)}} = 0$. Therefore, for hardware optimized GRBF KANs, the $RM_{edge_{(GRBF)}}$ can be recalculated as
\begin{align}
    RM_{edge_{RBF}} &= C_{fixed_{(GRBF)}} + C_{lin_{(GRBF)}} \notag \\
    &= N_c + 1.
\end{align}
Generalizing this relation to a dense KAN layer gives
\begin{equation}
    RM_{Layer_{(GRBF)}} = n_n n_i (N_c + 1).
\end{equation}

For the BOP calculation, the fixed overhead accounts for the base path multiplication which costs $b_ib_w$ BOPs. With LUT optimization, the GRBF evaluation incurs no BOPs since all basis function values are retrieved from memory. The linear combination $\sum_{i=0}^{N_c-1} w_i R_i(x)$ requires $N_c$ multiplications of weights of bitwidth $b_w$ with GRBF values of bitwidth $b_{rbf}$. This costs $N_c b_w b_{rbf}$ BOPs and $(N_c - 1)$ additions with accumulator bitwidth $\text{Acc}(N_c, b_w, b_{rbf})$. Thus, $BOP_{edge_{(GRBF)}}$ can be written as 
\begin{equation}
\begin{split}
    BOP_{edge_{(GRBF)}} = b_ib_w + N_c b_w b_{rbf} \\
    + (N_c - 1)\text{Acc}(N_c, b_w, b_{rbf}).
\end{split}
\end{equation}
For a dense GRBF-KAN layer, the total layer complexity includes both the per-edge costs across all $n_i \times n_n$ edges and the final output accumulation at each of the $n_n$ neurons. Therefore,
\begin{equation}
\begin{split}
    BOP_{Layer_{(GRBF)}} = n_n n_i [BOP_{edge_{(GRBF)}}] \\
    + n_n [(n_i - 1)(\text{Acc}(N_c, b_w, b_{rbf}) + \lceil \log_2 n_i \rceil)].
\end{split}
\end{equation}

The NABS calculation follows the same structure, replacing each multiplication with $X_w$ adders. The fixed overhead becomes $X_w(b_i + b_w)$ as the base path multiplication is replaced by $X_w$ adders. For the linear combination, the $N_c$ multiplications are each replaced by $X_w$ adders, and the $(N_c - 1)$ accumulation additions remain unchanged. Thus, $NABS_{edge_{(GRBF)}}$ are calculated as
\begin{equation}
\begin{split}
    NABS_{edge_{(GRBF)}} = X_w(b_i + b_w) \\
    + [N_c X_w + (N_c - 1)] \left[\text{Acc}(N_c, b_w, b_{rbf})\right].
\end{split}
\end{equation}
At the layer level, the NABS calculation accounts for all the edges and output accumulation. Thus,
\begin{equation}
\begin{split}
    NABS_{Layer_{(GRBF)}} = n_n n_i [NABS_{edge_{(GRBF)}}] \\
    + n_n [(n_i - 1)(\text{Acc}(N_c, b_w, b_{rbf}) + \lceil \log_2 n_i \rceil)].
\end{split}
\end{equation}

\begin{figure*}[t]
    \centering
    \includegraphics[width=0.95\textwidth]{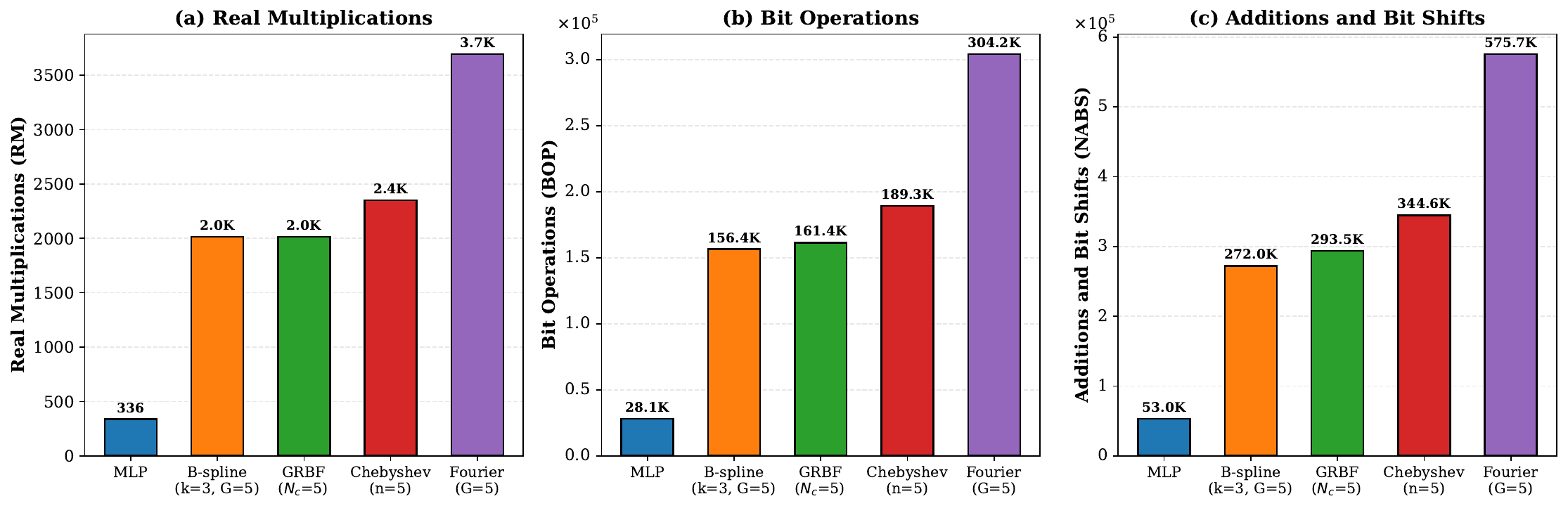}
    \caption{Comparison of hardware inference complexity for MLP and KAN variants using architecture $[3, 16, 16, 2]$. All KAN variants use representative parameters: B-spline ($k=3, G=5$), GRBF ($N_c=5$), Chebyshev ($n=5$), and Fourier ($G=5$). Results show (a) Real Multiplications (RM), (b) Bit Operations (BOP), and (c) Number of Additions and Bit Shifts (NABS).}
    
    \label{fig:kan_vs_mlp_1}
\end{figure*}

\subsection{Chebyshev KANs}

The real multiplications for Chebyshev KAN follow the same decomposition structure. The fixed overhead is once again a solitary multiplication for the base path $w_b \sigma(x)$. Unlike B-spline KAN, no input normalization multiplication is required since Chebyshev polynomials are defined on $[-1,1]$ and any input scaling can be absorbed into the LUT during precomputation. With LUT optimization, the $T_i(\tanh(x))$ composition can be precomputed and stored directly into $(n+1)$ global LUTs. Thus, all $(n+1)$ Chebyshev polynomial values are retrieved from memory without arithmetic operations. The linear combination $\sum_{i=0}^{n} w_i T_i(\tanh(x))$ requires $(n+1)$ multiplications. Therefore, for Chebyshev KANs, $(n+2)$ RMs are required per edge. For a dense layer, the total real multiplications $RM_{Layer_{(Cheby)}}$ are calculated as
\begin{equation}
    RM_{Layer_{(Cheby)}} = n_n n_i (n + 2).
\end{equation}

For the BOP calculation, the fixed overhead remains the same. The linear combination requires $(n+1)$ multiplications of $b_w$-bit weights with $b_{cheby}$-bit polynomial values and $n$ accumulations with accumulator bitwidth $\text{Acc}(n+1, b_w, b_{cheby})$. Therefore, the total BOP per edge and per layer for Chebyshev KANs can be calculated as
\begin{equation}
\begin{split}
    BOP_{edge_{(Cheby)}} = b_i b_w + (n+1) b_w b_{cheby} \\
    + n\left[\text{Acc}(n+1, b_w, b_{cheby})\right],
\end{split}
\end{equation}
\begin{equation}
\begin{split}
    BOP_{Layer_{(Cheby)}} = n_n n_i [BOP_{edge_{(Cheby)}}] \\
    + n_n [(n_i - 1)(\text{Acc}(n+1, b_w, b_{cheby}) + \lceil \log_2 n_i \rceil)].
\end{split}
\end{equation}

The NABS calculation follows naturally from the BOP calculation, as before, with multiplications replaced by adders. Hence, we derive:
\begin{equation}
\begin{split}
    NABS_{edge_{(Cheby)}} = X_w(b_i + b_w) \\
    + [(n+1)X_w + n]\left[\text{Acc}(n+1, b_w, b_{cheby})\right],
\end{split}
\end{equation}
\begin{equation}
\begin{split}
    NABS_{Layer_{(Cheby)}} = n_n n_i [NABS_{edge_{(Cheby)}}] \\
    + n_n [(n_i - 1)(\text{Acc}(n+1, b_w, b_{cheby}) + \lceil \log_2 n_i \rceil)].
\end{split}
\end{equation}

\subsection{Fourier KANs}

For Fourier KANs, all edges share the same set of $2G$ basis functions consisting of sine and cosine terms. Consistent with our treatment of exponentials in GRBF-KAN and the $\tanh$ transformation in Chebyshev KAN, we consider that these trigonometric functions are precomputed and stored in LUTs. The RM for Fourier KANs therefore consist of $1$ multiplication for the base path $w_b\sigma(x)$ and $2G$ multiplications for the linear combination path. Hence, the total RM number per edge is $(2G+1)$, and for a dense Fourier KAN layer, the total RM number can be calculated as
\begin{equation}
    RM_{Layer_{(Fourier)}} = n_n n_i (2G + 1).
\end{equation}

The BOP calculation follows the established procedure. The difference being that the linear combination requires $2G$ multiplications of $b_w$-bit weights with $b_{fourier}$-bit basis values and $(2G-1)$ accumulations. Thus, the per-edge and per-layer BOP calculations are given as
\begin{equation}
\begin{split}
    BOP_{edge_{(Fourier)}} = b_ib_w + (2G)b_w b_{fourier} \\
    + (2G-1)\text{Acc}(2G, b_w, b_{fourier}),
\end{split}
\end{equation}
\begin{equation}
\begin{split}
    BOP_{Layer_{(Fourier)}} = n_n n_i [BOP_{edge_{(Fourier)}}] \\
    + n_n [(n_i - 1)(\text{Acc}(2G, b_w, b_{fourier}) + \lceil \log_2 n_i \rceil)].
\end{split}
\end{equation}

Finally, the NABS calculation for Fourier KANs replaces the multiplications in the BOP calculation with adders giving
\begin{equation}
\begin{split}
    NABS_{edge_{(Fourier)}} = X_w(b_i + b_w) \\
    + [(2G)X_w + (2G-1)]\left[\text{Acc}(2G, b_w, b_{fourier})\right],
\end{split}
\end{equation}
\begin{equation}
\begin{split}
    NABS_{Layer_{(Fourier)}} = n_n n_i [NABS_{edge_{(Fourier)}}] \\
    + n_n [(n_i - 1)(\text{Acc}(2G, b_w, b_{fourier}) + \lceil \log_2 n_i \rceil)].
\end{split}
\end{equation}

\begin{figure*}[t]
    \centering
    \includegraphics[width=0.95\textwidth]{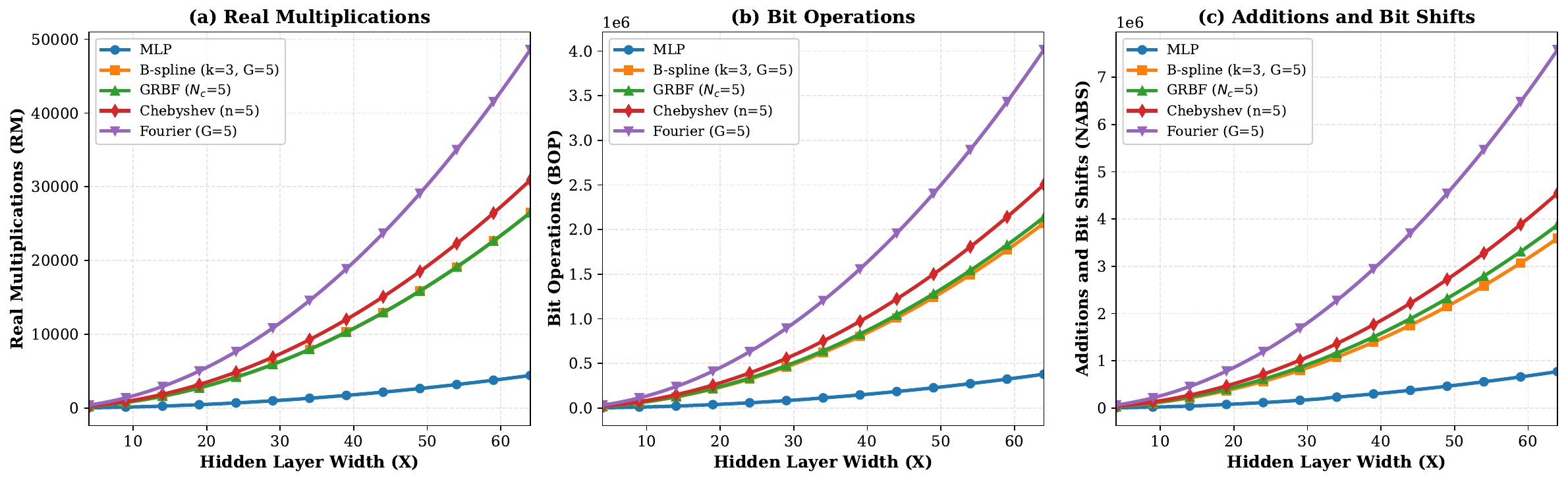}
    \caption{Complexity scaling with network width for architecture $[3, X, X, 2]$ where $X$ varies from 4 to 64. All networks exhibit quadratic scaling dominated by the $X \times X$ hidden layer. The computational overhead ratio between each KAN variant and MLP remains constant across network sizes.}
    \label{fig:kan_vs_mlp_2}
\end{figure*}

\begin{figure*}[t]
    \centering
    \includegraphics[width=0.95\textwidth]{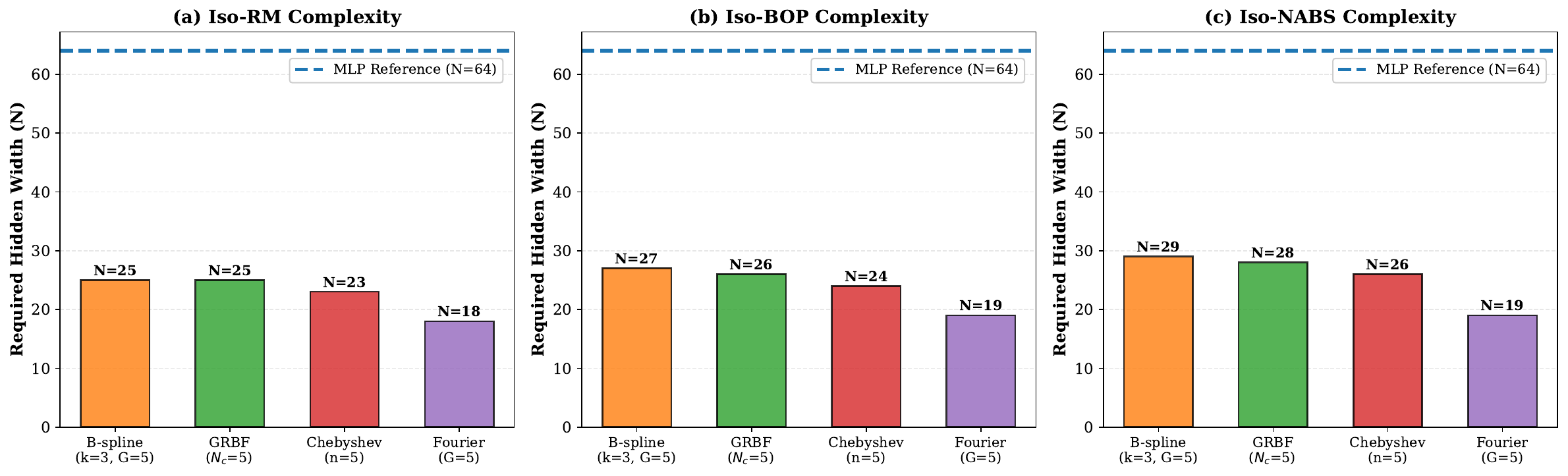}
    \caption{Iso-complexity analysis showing the required hidden layer width $X$ for KAN architecture $[3, X, X, 2]$ to match the computational cost of MLP baseline $[3, 64, 64, 2]$ across the three metrics i.e. (a) Real Multiplications (RM), (b) Bit Operations (BOP), and (c) Additions and Bit Shifts (NABS).}
    \label{fig:kan_vs_mlp_3}
\end{figure*}

\subsection{Cost of LUT-based Function Evaluation}
Throughout this paper, nonlinear functions are realized as LUTs. This assumption is applied uniformly to the base activation $\sigma(x)$ and to the basis functions of every KAN variant. A LUT-based evaluation is cost-free only when the table stores the outputs at full precision. In practice, platform-specific limits on table depth force a memory--arithmetic trade-off, in which case a smaller LUT is combined with on-the-fly interpolation to maintain accuracy. To make this explicit, we attach a uniform cost to each LUT-based evaluation, denoted as $\text{RM}_{\text{LUT}}$, $\text{BOP}_{\text{LUT}}$, and $\text{NABS}_{\text{LUT}}$. For a direct lookup using a full-precision table, $\text{RM}_{\text{LUT}}=\text{BOP}_{\text{LUT}}=\text{NABS}_{\text{LUT}}=0$. This zero-cost scenario recovers the closed-form results derived in Section~\ref{sec:derivation} and reported in Table~\ref{tab:complexity_all_networks}. For a linearly interpolated table utilizing a fractional bitwidth of $b_\delta$ and entry bitwidth of $b_{\text{tab}}$, the evaluation incurs the following computational costs:
\begin{align}
\label{eq:lut_cost}
    \text{RM}_{\text{LUT}} &= 1, \notag\\
    \text{BOP}_{\text{LUT}} &= b_\delta b_{\text{tab}} + \text{Acc}(2, b_\delta, b_{\text{tab}}), \notag\\
    \text{NABS}_{\text{LUT}} &= X_w(b_\delta + b_{\text{tab}}) + \text{Acc}(2, b_\delta, b_{\text{tab}}).
\end{align}
If $N_{\text{LUT}}$ denotes the number of LUT-based evaluations per edge, then the per-edge metrics generalize as
\begin{align}
\label{eq:rm_edge_lut}
    \text{RM}_{edge}^* &= \text{RM}_{edge} + N_{\text{LUT}}\,\text{RM}_{\text{LUT}}, \notag\\
    \text{BOP}_{edge}^* &= \text{BOP}_{edge} + N_{\text{LUT}}\,\text{BOP}_{\text{LUT}}, \notag\\
    \text{NABS}_{edge}^* &= \text{NABS}_{edge} + N_{\text{LUT}}\,\text{NABS}_{\text{LUT}}.
\end{align}
All numerical results in this paper assume the direct-lookup case and therefore represent the lower bound corresponding to full-precision tables. Implementing finite-precision tables shifts every architecture by the common additive factors defined in Eq.~\eqref{eq:lut_cost}.

\section{KANs VS MLP: Hypothetical Comparison in Terms of Proposed Metrics}
\label{sec:comparison}

Table~\ref{tab:complexity_all_networks} compares MLP and four KAN architectures in terms of the proposed hardware inference complexity metrics. To visualize these differences, we perform simulations considering uniform quantization with $8$-bit precision. The hardware complexity of KAN architectures depends on basis-specific parameters. For our comparative analysis, we select representative parameter values that balance computational cost with approximation capability. We use spline order of $k=3$ for B-splines, $N_c=5$ centers for GRBFs, a maximum degree of $n=5$ for Chebyshev polynomials and set the grid size to $G=5$ for Fourier KANs. These values were not selected via task-specific hyperparameter optimization. Instead, they were chosen to reflect configurations commonly reported in the KAN literature and to establish a structurally equivalent baseline across the evaluated KAN variants. A cubic spline serves as the de facto standard for B-spline KANs. The remaining variants were configured with a comparable number of active basis terms. Since the derived metrics are closed-form, the comparison can be generated for any other parameter set and the qualitative ordering of the variants is preserved as long as the active-term counts retain the same relation. It must be emphasized that these parameters are not universally optimal. In practice, parameter selection depends on multiple factors, such as problem complexity, computational budget, and accuracy requirements, or is determined by applying an optimization procedure~\cite{long2025genetic,buribayev2025darts}. These simulations serve to illustrate the fundamental computational characteristics of each KAN variant in terms of the metrics considered.

\subsection{Direct Architectural Comparison}

Figure~\ref{fig:kan_vs_mlp_1} compares hardware inference complexity for a fixed network architecture $[3, 16, 16, 2]$ that represents $3$ input features, $2$ hidden layers of $16$ neurons each, and $2$ output neurons. The results establish that KAN variants incur substantially higher computational costs than MLP for identical network width and depth. In terms of RM, a B-spline KAN with $k=3$ requires approximately $6\times$ more operations than an equivalent MLP. Similarly, B-spline KAN exhibits approximately $5.5\times$ higher BOPs and $5.1\times$ higher NABS as compared to MLP. However, studies have shown that KANs can achieve comparable accuracy to MLPs using smaller network architectures~\cite{somvanshi2025survey}. For instance, it was demonstrated that a $[17, 1, 14]$ KAN outperformed a $4$-layer, $300$ nodes wide MLP on a classification task~\cite{liu2025kan}. Similarly, in real-world satellite traffic forecasting, KANs achieved higher accuracy than MLPs using significantly fewer parameters~\cite{vaca2024kolmogorov}. Thus, enhanced learning capability can potentially offset the higher per-layer hardware complexity of the KAN architectures, making them a viable network choice for various problems. Moreover, since KANs break down complex multi-dimensional functions into a composition of simpler one-dimensional functions placed on network edges, they avoid the exponential growth in parameters required for high-dimensional mappings. For target functions with inherent compositional structure, this approach yields more favorable scaling between parameter count and approximation error as compared to standard MLPs~\cite{zhang2025generalization}.

Among KAN variants, complexity correlates with the number of active basis functions. Since basis function evaluation in hardware is LUT-based, the computational cost arises entirely from the linear combination of active terms. With the chosen parameters, for B-spline KANs, only $(k+1)=4$ terms are active for a given input, regardless of the grid size. In contrast, GRBF, Chebyshev and Fourier KANs compute weighted sums over all terms, i.e. $N_c=5$, $(n+1)=6$, and $2G=10$ terms, respectively. Thus, we observe lower complexity for the B-spline KANs as compared to the other three KAN variants with the chosen parameters.

\subsection{Network Scaling Analysis}

Figure~\ref{fig:kan_vs_mlp_2} examines how the complexity scales with network width. For the architecture $[3, X, X, 2]$, $X$ is varied from $4$ to $64$. All networks exhibit quadratic scaling dominated by the $X^2$ term from the second hidden layer. However, the rate of growth differs according to the per-edge overhead of each architecture. For example, B-spline KAN maintains the constant $6\times$ overhead in terms of RM as compared to MLP observed in the direct comparison. Thus, the computational overhead ratio between KAN and MLP remains constant across network sizes.

\subsection{Iso-Complexity Analysis}

Since KANs have a higher cost per edge, direct layer-for-layer comparisons bias the results against KANs. Therefore, a more meaningful comparison examines architectural equivalence from a computational budget perspective. To investigate the latter, we analyze which KAN network configurations consume the same resources as compared to an MLP baseline. Figure~\ref{fig:kan_vs_mlp_3} illustrates an iso-complexity analysis, where we sweep the hidden layer size $X$ of $[3, X, X, 2]$ KANs and compare this against a $[3, 64, 64, 2]$ MLP. As expected from the previous direct complexity evaluations, the results show that B-spline KAN achieves the widest networks within the target complexity budget, with $N \approx 25$-$29$ depending on the metric. Since Fourier KANs have the highest number of active basis terms that need to be evaluated for the chosen parameter settings, they exhibit the tightest constraints, allowing only $N \approx 18$-$19$. GRBF and Chebyshev KANs occupy intermediate positions. 

Moreover, an important validation of our complexity framework emerges from the consistency across metrics. For each KAN variant, the required hidden-layer width differs only slightly between RM, BOP, and NABS. Despite measuring fundamentally different quantities, all three metrics converge on nearly the same architectural equivalences. This tight clustering indicates that the mathematical formulations correctly capture the computational costs. Overall, our results highlight and quantify the fundamental trade-off that for a given computational budget, KANs must use narrower networks as compared to MLPs. A complete interpretation of this result must also consider accuracy. The iso-complexity analysis presented here establishes only that a KAN must have narrower width to fit within an MLP's hardware operations budget. It does not guarantee that a narrower KAN retains the reference accuracy. Recent studies report that KANs can match or exceed MLP performance utilizing smaller network sizes~\cite{liu2025kan, vaca2024kolmogorov, somvanshi2025survey, baravsin2025exploring}. However, whether the specific iso-complexity widths reported in this study successfully preserve accuracy is entirely task-dependent. We therefore present these results as the budgets within which a KAN must operate to remain competitive with the chosen MLP baseline and consider a joint accuracy--complexity study across representative tasks as necessary future work. 

\section{Scope and Limitations}
The proposed analytical metrics deliberately isolate the arithmetic cost of inference, which maintains their platform independence. However, this leaves certain practical factors outside their scope. First, the RM, BOP, and NABS metrics exclusively count computational operations. They do not model data movement or memory-bandwidth costs. For large networks, the energy and latency required to move weights and LUT contents between memory and compute units can dominate the arithmetic cost. Consequently, the choice between on-chip and off-chip storage becomes critical and must be evaluated independently during physical hardware synthesis. When all network weights and shared basis LUTs fit within on-chip memory, the metrics closely track the total inference cost. However, once the parameter count exceeds on-chip capacity and spills into off-chip memory, bandwidth becomes the primary bottleneck for system throughput. Since each KAN edge requires multiple coefficients as compared to a single weight for an MLP edge, KANs will exhaust on-chip memory and enter the bandwidth-limited off-chip memory regime at smaller network widths than MLPs. This memory-bound scaling behavior is a physical implementation constraint that the proposed metrics do not capture.

Second, the metrics do not express latency or throughput. Two designs with identical operation counts can yield different latency and throughput figures depending on how the the clock frequency, initiation interval. and pipeline depth are fixed. Third, the proposed analytical framework has not been validated against physical FPGA or ASIC logic synthesis. Our objective is to evaluate the pure algorithmic complexity of KAN architectures prior to hardware implementation since post-synthesis metrics are inherently platform-dependent. The analytical metrics derived in this work provide a platform-independent baseline that enables rapid design-space exploration across various KAN topologies without incurring the time cost of physical synthesis. Finally, our analysis covers four representative basis families chosen to span the dominant design paradigms. The computational complexity of other KAN variants, such as Fractional KANs~\cite{aghaei2025fkan} and Wavelet KANs~\cite{bozorgasl2024wavkanwaveletkolmogorovarnoldnetworks, seydi2024unveiling}, can be evaluated by following the same edge decomposition methodology. To maintain clarity, we restrict the present comparison to the four most widely benchmarked families of basis functions.

\section{Conclusion}
\label{sec:conclusion}
In this paper, we establish a comprehensive analytical framework for evaluating the hardware-oriented inference complexity of KANs. Existing complexity assessments based on FLOPs provide appropriate estimates of complexity for GPU-based training and inference, but fail to capture the true cost of specialized hardware implementations. Moreover, existing hardware studies for KANs report complexity in terms of platform-specific resource consumption metrics that are not suitable for early-stage design space exploration. To address these limitations, we derive generalized formulae for evaluating complexity in terms of RM, BOP, and NABS. To the best of our knowledge, this is the first study to comprehensively quantify these metrics across four distinct KAN architectures, i.e., B-spline, GRBF, Chebyshev, and Fourier KANs. Importantly, the results presented in this paper can be readily extended to other KAN architectures as well. We also note that these metrics quantify arithmetic operations only. They do not capture data movement and memory bandwidth costs, on-chip versus off-chip storage effects, or the impact on latency and throughput. Moreover, we regard a joint accuracy--complexity evaluation and a direct comparison against post-synthesis resource utilization as important avenues for future work. In addition, an optimized and task-specific comparison of KANs with other modern neural network architectures in terms of these metrics is another promising research direction.

Our findings affirm that a KAN network incurs a higher computational cost as compared to an identically sized MLP network. To operate under a fixed complexity budget, KANs must use smaller network sizes as compared to MLP. The platform-independent complexity metrics presented in this paper equip the research community with a robust and fair framework for evaluating the inference complexity of KANs. We believe that comparing KANs with other DNNs in terms of these metrics across diverse tasks will further enhance the understanding of KANs' practical viability in engineering applications.

\section*{Acknowledgment}\label{sec:ack}
This research was supported by the European Union's Horizon Europe research and innovation programme MSCA-DN NESTOR (G.A. 101119983), and by UK Research and Innovation (UKRI) under the Horizon Europe Guarantee scheme (Grant Ref: EP/Y031024/1). Sergei K. Turitsyn also acknowledges EPSRC project TRANSNET (EP/R035342/1). Experiments were run on Aston EPS Machine Learning Server, funded by the EPSRC Core Equipment Fund, Grant EP/V036106/1. For the purposes of open access, the authors have applied a Creative Commons Attribution (CC BY) license to any Author Accepted Manuscript (AAM) version arising from this submission.

\bibliographystyle{IEEEtran}
\bibliography{references}

\end{document}